\documentclass{article}

\PassOptionsToPackage{numbers}{natbib}
\usepackage[final]{nips_2016}

\usepackage[utf8]{inputenc} 
\usepackage[T1]{fontenc}    
\usepackage{hyperref}       
\usepackage{url}            
\usepackage{booktabs}       
\usepackage{amsfonts}       
\usepackage{nicefrac}       
\usepackage{microtype}      
\usepackage{graphicx}
\usepackage{subfig}
\usepackage{color}
\usepackage{multirow}
\usepackage{array}
\usepackage{amsopn}
\usepackage{wrapfig}

\def\R{\mathbb{R}}

\DeclareMathOperator*{\logadd}{logadd}

\title{Wav2Letter: an End-to-End ConvNet-based Speech Recognition System}

\author{
    Ronan Collobert\\
    Facebook AI Research, Menlo Park\\
    \texttt{locronan@fb.com}\\
    \And
    Christian Puhrsch\\
    Facebook AI Research, Menlo Park\\
    \texttt{cpuhrsch@fb.com}\\
    \And
    Gabriel Synnaeve\\
    Facebook AI Research, New York\\
    \texttt{gab@fb.com}\\
}

\begin{document}

\maketitle

\begin{abstract}
    This paper presents a simple end-to-end model for speech recognition,
    combining a convolutional network based acoustic model and a graph
    decoding. It is trained to output letters, with transcribed speech,
    without the need for force alignment of phonemes. We introduce an
    automatic segmentation criterion for training from sequence annotation
    without alignment that is on par with CTC
    \citep{graves2006connectionist} while being simpler. We show
    competitive results in word error rate on the Librispeech corpus
    \citep{panayotov2015librispeech} with MFCC features, and promising
    results from raw waveform.
\end{abstract}

\section{Introduction}

We present an end-to-end system to speech recognition, going from the speech
signal (e.g. Mel-Frequency Cepstral Coefficients (MFCC), power spectrum, or raw
waveform) to the transcription. The acoustic model is trained using letters
(graphemes) directly, which take out the need for an intermediate (human or
automatic) phonetic transcription. Indeed, the classical pipeline to build
state of the art systems for speech recognition consists in first training an
HMM/GMM model to force align the units on which the final acoustic model
operates (most often context-dependent phone states). This approach takes its
roots in HMM/GMM training \citep{woodland1993htk}. The improvements brought by
deep neural networks (DNNs) \citep{mohamed2012acoustic,hinton2012deep} and
convolutional neural networks (CNNs) \citep{sercu2015very,soltau2014joint} for
acoustic modeling only extend this training pipeline.

The current state of the art on Librispeech (the dataset that we used for our
evaluations) uses this approach too
\citep{panayotov2015librispeech,peddinti2015time}, with an additional step of
speaker adaptation \citep{saon2013speaker,peddinti2015jhu}. Recently,
\cite{senior2014gmm} proposed GMM-free training, but the approach still
requires to generate a force alignment. An approach that cut ties with the
HMM/GMM pipeline (and with force alignment) was to train with a recurrent
neural network (RNN) \citep{graves2013speech} for phoneme transcription.  There
are now competitive end-to-end approaches of acoustic models toppled with RNNs
layers as in \citep{hannun2014deep,miao2015eesen,saon2015ibm,amodei2015deep}, trained with a
sequence criterion \citep{graves2006connectionist}. However these models are
computationally expensive, and thus take a long time to train. 

Compared to classical approaches that need phonetic annotation (often derived
from a phonetic dictionary, rules, and generative training), we propose to
train the model end-to-end, using graphemes directly. Compared to sequence
criterion based approaches that train directly from speech signal to graphemes
\citep{miao2015eesen}, we propose a simple(r) architecture (23 millions of
parameters for our best model, vs. 100 millions of parameters in
\citep{amodei2015deep}) based on convolutional networks for the acoustic model,
toppled with a graph transformer network \citep{bottou1997global}, trained with
a simpler sequence criterion.  Our word-error-rate on clean speech is slightly
better than \citep{hannun2014deep}, and slightly worse than
\citep{amodei2015deep}, in particular factoring that they train on 12,000 hours
while we only train on the 960h available in LibriSpeech's train set. Finally,
some of our models are also trained on the raw waveform, as in
\citep{palaz2013estimating,palaz2015analysis}.
The rest of the paper is structured as follows: the next section presents the
convolutional networks used for acoustic modeling, along with the automatic
segmentation criterion. The following section shows experimental results
comparing different features, the criterion, and our current best word error
rates on LibriSpeech.


\section{Architecture}

Our speech recognition system is a standard convolutional neural
network \citep{lecun1995convolutional} fed with various different features,
trained through an alternative to the Connectionist Temporal Classification
(CTC) \citep{graves2006connectionist}, and coupled with a simple beam search
decoder. In the following sub-sections, we detail each of these components.

\subsection{Features}

\begin{wrapfigure}[40]{r}{4cm}
  \centering
  \includegraphics[height=12cm]{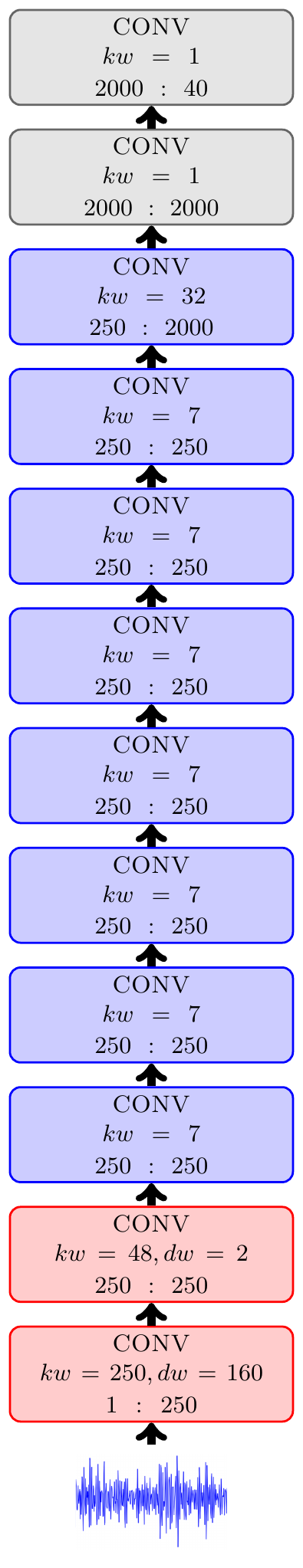}
  \caption{
    \label{fig-raw-arch}
    Our neural network architecture for raw wave. First two layers
    are convolutions with strides. Last two layers are convolutions with
    $kw=1$, which are equivalent to fully connected layers. Power spectrum
    and MFCC based networks do not have the first layer.  }
\end{wrapfigure}

We consider three types of input features for our model: MFCCs, power-spectrum,
and raw wave. MFCCs are carefully designed speech-specific features, often
found in classical HMM/GMM speech systems \citep{woodland1993htk} because of
their dimensionality compression (13 coefficients are often enough to span
speech frequencies). Power-spectrum features are found in most recent deep
learning acoustic modeling features \citep{amodei2015deep}. Raw wave has been
somewhat explored in few recent work
\citep{palaz2013estimating,palaz2015analysis}. ConvNets have the advantage to
be flexible enough to be used with either of these input feature types.  Our
acoustic models output letter scores (one score per letter, given a dictionary ${\cal L}$).

\subsection{ConvNet Acoustic Model}

The acoustic models we considered in this paper are all based on standard 1D
convolutional neural networks (ConvNets). ConvNets interleave convolution
operations with pointwise non-linearity operations. Often ConvNets also
embark pooling layers: these type of layers allow the network to ``see'' a
larger context, without increasing the number of parameters, by locally
aggregating the previous convolution operation output. Instead, our
networks leverage striding convolutions. Given $(x_t)_{t=1\dots T_x}$ an
input sequence with $T_x$ frames of $d_x$ dimensional vectors, a
convolution with kernel width $kw$, stride $dw$ and $d_y$ frame size output
computes the following:
\begin{equation}
y_t^i = b_i + \sum_{j=1}^{d_x} \sum_{k=1}^{kw} w_{i,j,k}\, x^j_{dw\times (t-1)+k}\,\quad\forall 1\leq i\leq d_y,
\end{equation}
where $b\in\R^{d_y}$ and $w\in\R^{d_y\times d_x\times kw}$ are the parameters
of the convolution (to be learned).

Pointwise non-linear layers are added after convolutional layers. In our
experience, we surprisingly found that using hyperbolic tangents, their
piecewise linear counterpart HardTanh (as in \citep{palaz2015analysis}) or ReLU
units lead to similar results.

There are some slight variations between the architectures, depending on
the input features.  MFCC-based networks need less striding, as
standard MFCC filters are applied with large strides on the input raw
sequence. With power spectrum-based and raw wave-based networks, we
observed that the overall stride of the network was more
important than where the convolution with strides were placed. We found thus preferrable to
set the strided convolutions near the first input layers of the network, as it leads to
the fastest architectures: with power spectrum features or raw wave, the
input sequences are very long and the first convolutions are thus the most
expensive ones.

The last layer of our convolutional network outputs one score per letter in the
letter dictionary ($d_y=|{\cal L}|$). Our architecture for raw wave is shown in
Figure~\ref{fig-raw-arch} and is inspired by \citep{palaz2015analysis}.  The
architectures for both power spectrum and MFCC features do not include the
first layer. The full network can be seen as a non-linear convolution, with a
kernel width of size $31280$ and stride equal to $320$; given the sample rate
of our data is 16KHz, label scores are produced using a window of 1955 ms,
with steps of 20ms.

\subsection{Inferring Segmentation with AutoSegCriterion}
\label{sec-autosegcriterion}

Most large labeled speech databases provide only a text transcription for
each audio file. In a classification framework (and given our acoustic
model produces letter predictions), one would need the segmentation of each
letter in the transcription to train properly the model. Unfortunately,
manually labeling the segmentation of each letter would be tedious. Several
solutions have been explored in the speech community to alleviate this
issue: HMM/GMM models use an iterative EM procedure: (i) during the
Estimation step, the best segmentation is inferred, according to the
current model, by maximizing the joint probability of the letter (or any
sub-word unit) transcription and input sequence. (ii) During the
Maximization step the model is optimized by minimizing a frame-level
criterion, based on the (now fixed) inferred segmentation. This approach is
also often used to boostrap the training of neural network-based acoustic
models.

Other alternatives have been explored in the context of hybrid HMM/NN
systems, such as the MMI criterion~\citep{bahl1986icassp} which maximizes the
mutual information between the acoustic sequence and word sequences or the
Minimum Bayse Risk (MBR) criterion~\citep{gibson2006interspeech}.

\begin{figure}
  \centering
  \subfloat[]{\label{fig-ctc-g}\includegraphics[width=0.45\linewidth]{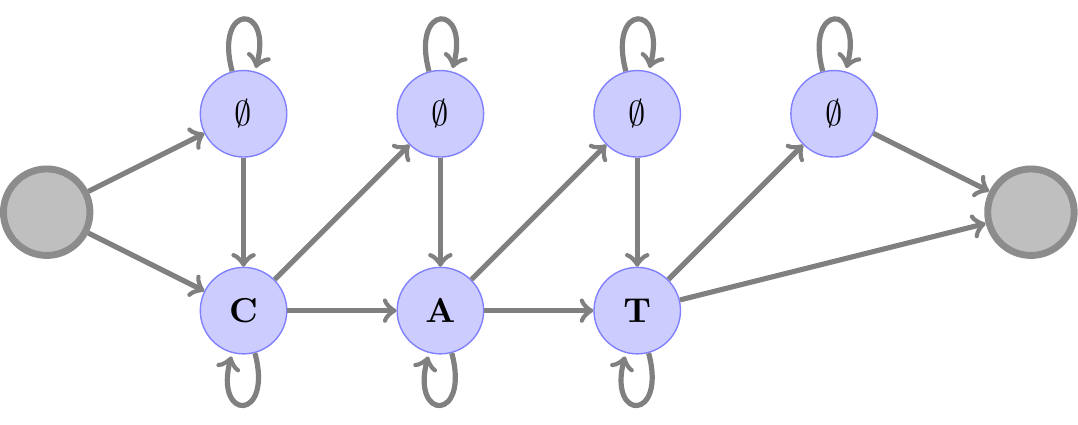}}
  \hspace*{0.5cm}
  \subfloat[]{\label{fig-ctc-ug}\includegraphics[width=0.45\linewidth]{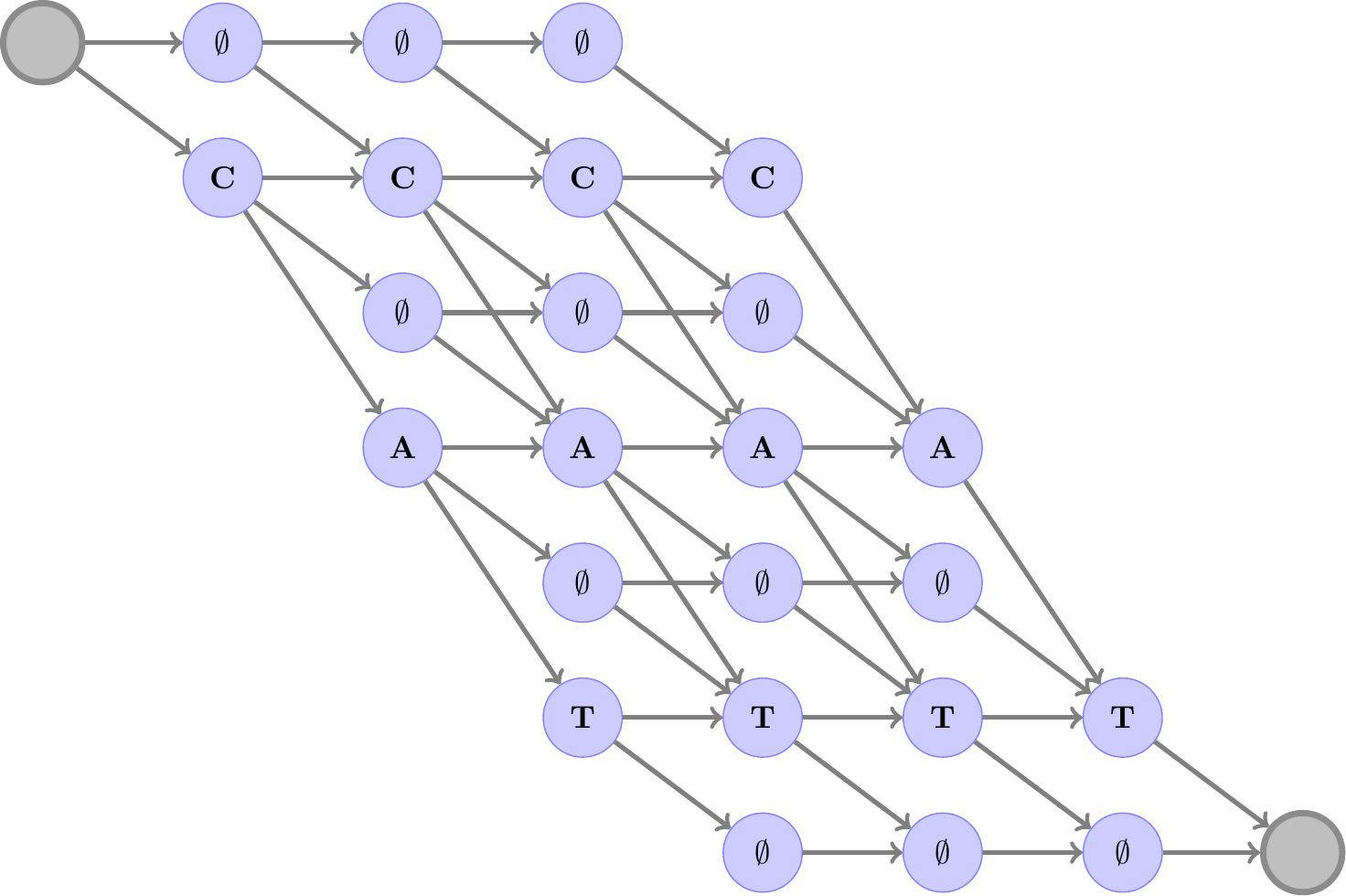}}
  \caption{
    \label{fig-ctc} The CTC criterion graph. (a) Graph which represents all the acceptable sequences
    of letters (with the blank state denoted ``$\emptyset$''), for the
    transcription ``cat''. (b) Shows the same graph unfolded over 5
    frames. There are no transitions scores. At each time step, nodes are
    assigned a conditional probability output by the neural network
    acoustic model.}
\end{figure}

More recently, standalone neural network architectures have been trained
using criterions which jointly infer the segmentation of the transcription
while increase the overall score of the right
transcription~\citep{graves2006connectionist,palaz2014joint}. The most
popular one is certainly the Connectionist Temporal Classification (CTC)
criterion, which is at the core of Baidu's Deep Speech
architecture~\citep{amodei2015deep}. CTC assumes that the network output
probability scores, normalized at the frame level. It considers all
possible sequence of letters (or any sub-word units), which can lead to a
to a given transcription. CTC also allow a special ``blank'' state to be
optionally inserted between each letters. The rational behind the blank
state is two-folds: (i) modeling ``garbage'' frames which might occur
between each letter and (ii) identifying the separation between two
identical consecutive letters in a transcription. Figure~\ref{fig-ctc-g}
shows an example of the sequences accepted by CTC for a given
transcription. In practice, this graph is unfolded as shown in
Figure~\ref{fig-ctc-ug}, over the available frames output by the acoustic
model. We denote ${\cal G}_{ctc}(\theta, T)$ an unfolded graph over $T$
frames for a given transcription $\theta$, and $\pi = {\pi_1,\,\dots,\,
  \pi_T} \in {\cal G}_{ctc}(\theta, T)$ a path in this graph representing a
(valid) sequence of letters for this transcription. At each time step $t$,
each node of the graph is assigned with the corresponding log-probability
letter (that we denote $f_t(\cdot)$) output by the acoustic model. CTC aims at
maximizing the ``overall'' score of paths in ${\cal G}_{ctc}(\theta, T)$;
for that purpose, it minimizes the Forward score:
\begin{equation}
  CTC(\theta, T) = - \logadd_{\pi \in {\cal G}_{ctc}(\theta, T)} \sum_{t=1}^T f_{\pi_t}(x)\,,
  \label{eq-ctc}
\end{equation}
where the ``logadd'' operation, also often called ``log-sum-exp'' is
defined as $\logadd(a, b) = \exp(\log(a) + \log(b))$. This overall score
can be efficiently computed with the Forward algorithm. To put things in
perspective, if one would replace the $\logadd(\cdot)$ by a $\max(\cdot)$
in~(\ref{eq-ctc}) (which can be then efficiently computed by the Viterbi
algorithm, the counterpart of the Forward algorithm), one would then
maximize the score of the \emph{best} path, according to the model
belief. The $\logadd(\cdot)$ can be seen as a smooth version of the
$\max(\cdot)$: paths with similar scores will be attributed the same weight
in the overall score (and hence receive the same gradient), and paths with
much larger score will have much more overall weight than paths with low
scores. In practice, using the $\logadd(\cdot)$ works much better than the
$\max(\cdot)$. It is also worth noting that maximizing~(\ref{eq-ctc}) does
not diverge, as the acoustic model is assumed to output normalized scores
(log-probabilities) $f_i(\cdot)$.

\begin{figure}
  \centering
  \subfloat[]{\label{fig-asg-g}\includegraphics[width=0.45\linewidth]{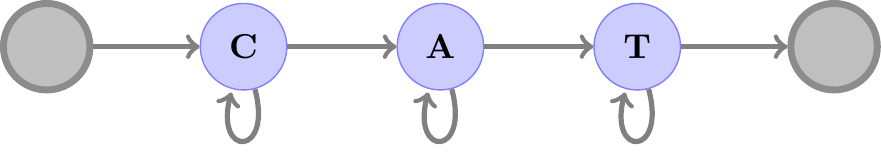}}
  \hspace*{0.5cm}
  \subfloat[]{\label{fig-asg-ug}\includegraphics[width=0.45\linewidth]{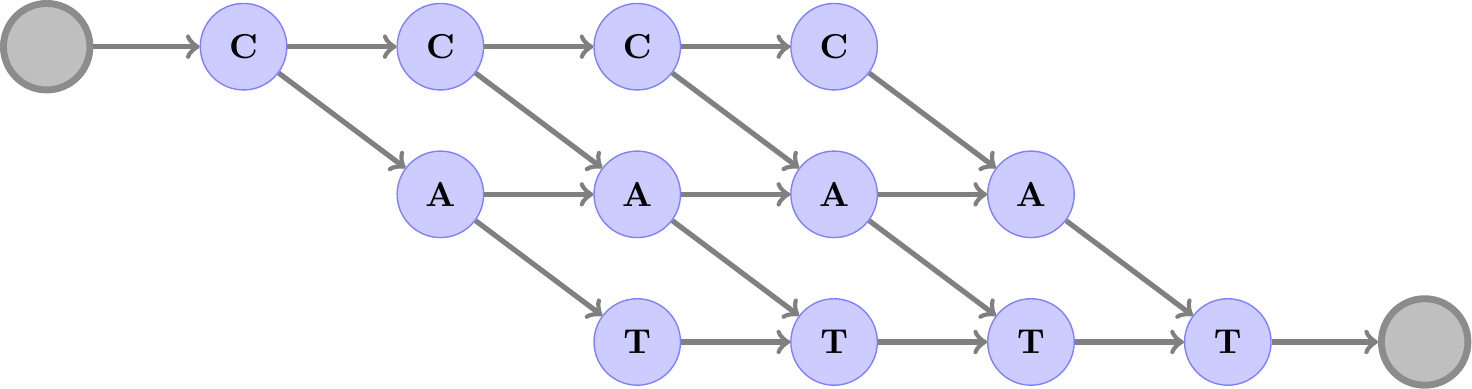}}\\
  \subfloat[]{\label{fig-full-ug}\includegraphics[width=0.45\linewidth]{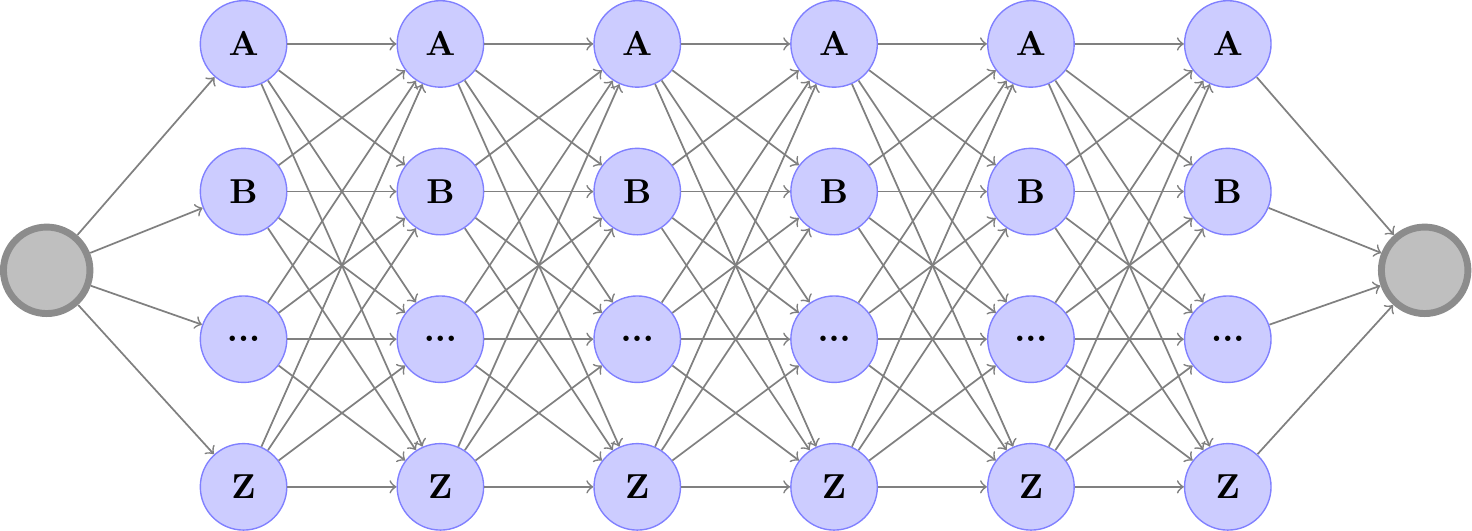}}
  \caption{
    \label{fig-asg} The ASG criterion graph.
    (a) Graph which represents all the acceptable sequences
    of letters for the transcription ``cat''. (b) Shows the same graph
    unfolded over 5 frames. (c) Shows the corresponding fully connected
    graph, which describe all possible sequences of letter; this graph is
    used for normalization purposes. Un-normalized transitions scores are
    possible on the edges. At each time step, nodes are assigned a
    conditional un-normalized score, output by the neural network acoustic
    model.}
\end{figure}

In this paper, we explore an alternative to CTC, with three differences:
(i)~there are no blank labels, (ii)~un-normalized scores on the nodes (and possibly
un-normalized transition scores on the edges) (iii)~global normalization
instead of per-frame normalization:
\begin{itemize}
\item The advantage of (i) is that it produces a much simpler graph (see
  Figure~{\ref{fig-asg-g}} and Figure~{\ref{fig-asg-ug}}). We found that in
  practice there was no advantage of having a blank class to model the
  possible ``garbage'' frames between letters. Modeling letter repetitions
  (which is also an important quality of the blank label in CTC) can be
  easily replaced by repetition character labels (we used two extra labels
  for two and three repetitions). For example ``caterpillar'' could be
  written as ``caterpil2ar'', where ``2'' is a label to represent the
  repetition of the previous letter. Not having blank labels also
  simplifies the decoder.
\item With (ii) one can easily plug an external language model, which would
  insert transition scores on the edges of the graph. This could be
  particularly useful in future work, if one wanted to model
  representations more high-level than letters. In that respect, avoiding
  normalized transitions is important to alleviate the problem of ``label
  bias''~\cite{leon1991approche,lafferty2001crf}. In this work, we limited
  ourselves to transition scalars, which are learned together with the
  acoustic model.
  \item The normalization evoked in~(iii) is necessary when using
    un-normalized scores on nodes or edges; it insures incorrect
    transcriptions will have a low confidence.
\end{itemize}
In the following, we name our criterion ``Auto Segmentation Criterion''
(ASG). Considering the same notations than for CTC in~(\ref{eq-ctc}), and
an unfolded graph ${\cal G}_{asg}(\theta, T)$ over $T$ frames for a given
transcription $\theta$ (as in Figure~\ref{fig-asg-ug}), as well as a fully
connected graph ${\cal G}_{full}(\theta, T)$ over $T$ frames (representing
all possible sequence of letters, as in Figure~\ref{fig-full-ug}), ASG aims at minimizing:
\begin{equation}
  ASG(\theta, T) = - \logadd_{\pi \in {\cal G}_{asg}(\theta, T)} \sum_{t=1}^T (f_{\pi_t}(x)+g_{\pi_{t-1},\pi_t}(x)) + \logadd_{\pi \in {\cal G}_{full}(\theta, T)} \sum_{t=1}^T (f_{\pi_t}(x)+g_{\pi_{t-1},\pi_t}(x)) \,,
  \label{eq-asg}
\end{equation}
where $g_{i,j}(\cdot)$ is a transition score model to jump from label $i$
to label $j$.  The left-hand part of $\ref{eq-asg}$ promotes sequences of
letters leading to the right transcription, and the right-hand part demotes
all sequences of letters. As for CTC, these two parts can be efficiently
computed with the Forward algorithm. Derivatives with respect to
$f_i(\cdot)$ and $g_{i,j}(\cdot)$ can be obtained (maths are a bit tedious)
by applying the chain rule through the Forward recursion.

\subsection{Beam-Search Decoder}
\label{sec-decoder}
We wrote our own one-pass decoder, which performs a simple beam-search with
beam threholding, histogram pruning and language model
smearing~\cite{steinbiss1994improvements}. We kept the decoder as simple as
possible (under 1000 lines of C code). We did not implement any sort of
model adaptation before decoding, nor any word graph rescoring. Our decoder
relies on KenLM~\cite{heafield2013scalable} for the language modeling
part. It also accepts un-normalized acoustic scores (transitions and
emissions from the acoustic model) as input. The decoder attempts to maximize
the following:
\begin{equation}
  \label{eq-decoder}
{\cal L}(\theta) =  \logadd_{\pi \in {\cal G}_{asg}(\theta, T)} \sum_{t=1}^T (f_{\pi_t}(x)+g_{\pi_{t-1},\pi_t}(x)) + \alpha \log P_{lm}(\theta) + \beta |\theta|\,,
\end{equation}
where $P_{lm}(\theta)$ is the probability of the language model given a
transcription $\theta$, $\alpha$ and $\beta$ are two hyper-parameters which
control the weight of the language model and the word insertion penalty
respectively.

\section{Experiments}


We implemented everything using
Torch7\footnote{\url{http://www.torch.ch}.}. The ASG criterion as well as
the decoder were implemented in C (and then interfaced into Torch).

We consider as benchmark LibriSpeech, a large speech database freely available
for download~\citep{panayotov2015librispeech}. LibriSpeech comes with its own
train, validation and test sets. Except when specified, we used all the
available data (about 1000h of audio files) for training and validating our
models. We use the original 16 KHz sampling rate. The vocabulary ${\cal L}$
contains 30 graphemes: the standard English alphabet plus the apostrophe,
silence, and two special ``repetition'' graphemes which encode the duplication
(once or twice) of the previous letter (see
Section~\ref{sec-autosegcriterion}).

The architecture hyper-parameters, as well the decoder ones were tuned
using the validation set. In the following, we either report
letter-error-rates (LERs) or word-error-rates (WERs). WERs have been
obtained by using our own decoder (see Section~\ref{sec-decoder}), with the
standard 4-gram language model provided with
LibriSpeech\footnote{\url{http://www.openslr.org/11}.}.

MFCC features are computed with 13 coefficients, a 25 ms sliding window and
10 ms stride. We included first and second order derivatives. Power
spectrum features are computed with a 25 ms window, 10 ms stride, and have
257 components. All features are normalized (mean 0, std 1) per input sequence.



\subsection{Results}



Table~\ref{tbl-asg-vs-ctc} reports a comparison between CTC and ASG, in terms
of LER and speed. Our ASG criterion is implemented in C (CPU only), leveraging
SSE instructions when possible. Our batching is done with an OpenMP parallel
for. We picked the CTC criterion implementation provided by
Baidu\footnote{\url{https://github.com/baidu-research/warp-ctc}.}. Both
criteria lead to the same LER. For comparing the speed, we report performance
for sequence sizes as reported initially by Baidu, but also for longer sequence
sizes, which corresponds to our average use case.  ASG appears faster on long
sequences, even though it is running on CPU only. Baidu's GPU CTC
implementation seems more aimed at larger vocabularies (e.g. 5000 Chinese
characters).
\begin{table}
  \caption{CTC vs ASG. CTC is Baidu's implementation. ASG is implemented on
    CPU (core in C, threading in Lua). (a) reports performance in
    LER. Timings (in \emph{ms}) for small sequences (input frames: 150,
    letter vocabulary size: 28, transcription size: 40) and long sequences
    (input frames: 700, letter vocabulary size: 28, transcription size:
    200) are reported in (b) and (c) respectively. Timings include both
    forward and backward passes. CPU implementations use 8 threads.
\label{tbl-asg-vs-ctc}
  }
  \centering
  \subfloat[]{\label{tbl-asg-vs-ctc-ler}
    \begin{tabular}{|l|c|c|}
      \hline
      & ASG & CTC \\
      \hline
      dev-clean & 10.4 & 10.7 \\
      test-clean & 10.1 & 10.5 \\
      \hline
    \end{tabular}
  }
  \
  \subfloat[]{\label{fig-asg-vs-ctc-speed-small}
    \begin{tabular}{|c|r|r|r|}
      \hline
      batch & \multicolumn{2}{|c|}{CTC} & ASG \\
      size  & CPU & GPU & CPU \\
      \hline
      1 & 1.9 & 5.9 & 2.5 \\
      4 & 2.0 & 6.0 & 2.8 \\
      8 & 2.0 & 6.1 & 2.8 \\
      \hline
    \end{tabular}
  }
  \\
  \subfloat[]{\label{fig-asg-vs-ctc-speed-large}
    \begin{tabular}{|c|r|r|r|}
      \hline
      batch & \multicolumn{2}{|c|}{CTC} & ASG \\
      size  & CPU & GPU & CPU \\
      \hline
      1 & 40.9 & 97.9 & 16.0 \\
      4 & 41.6 & 99.6 & 17.7 \\
      8 & 41.7 & 100.3 & 19.2 \\
      \hline
    \end{tabular}
  }
\end{table}

\begin{figure}
  \centering
  \subfloat[]{\label{fig-ler-vs-h}\includegraphics[width=0.45\linewidth]{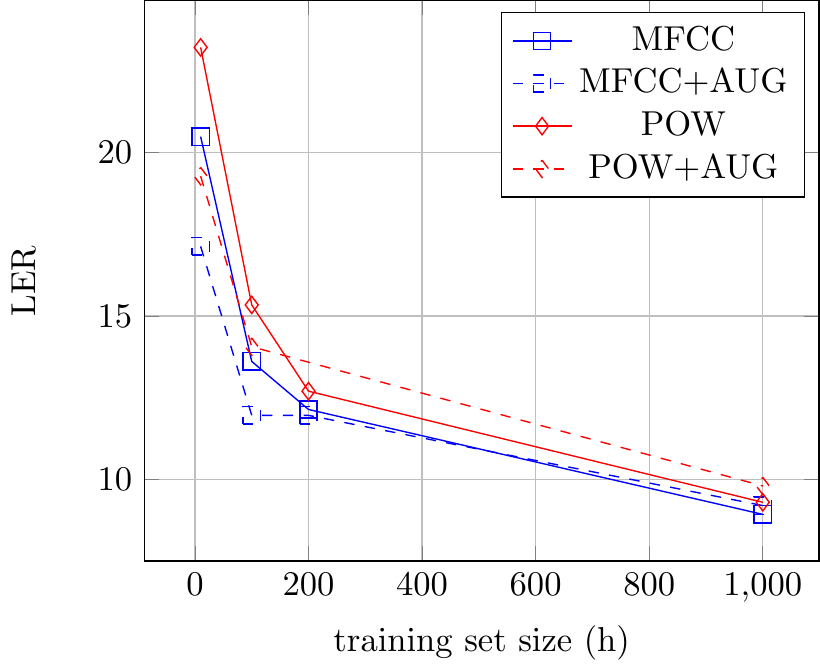}}
  \hspace*{0.5cm}
  \subfloat[]{\label{fig-wer-vs-h}\includegraphics[width=0.45\linewidth]{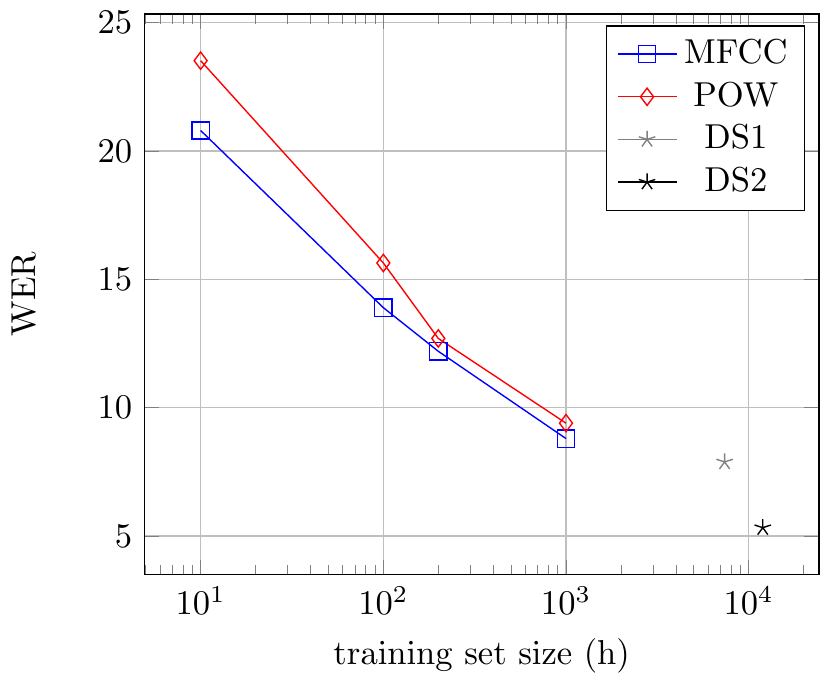}}
  \caption{\label{fig-vs-h} Valid LER (a) and WER (b) v.s. training set size (10h, 100h,
    200h, 1000h). This compares MFCC-based and power spectrum-based (POW)
    architectures.  AUG experiments include data augmentation. In (b) we
    provide Baidu Deep Speech 1 and 2 numbers on LibriSpeech, as a
    comparison~\cite{hannun2014deep,amodei2015deep}.  }
\end{figure}

We also investigated the impact of the training size on the dataset, as
well as the effect of a simple data augmentation procedure, where shifts
were introduced in the input frames, as well as stretching. For that
purpose, we tuned the size of our architectures (given a particular size of
the dataset), to avoid over-fitting. Figure~\ref{fig-ler-vs-h} shows the
augmentation helps for small training set size. However, with enough
training data, the effect of data augmentation vanishes, and both type of
features appear to perform similarly. Figure~\ref{fig-wer-vs-h} reports the
WER with respect to the available training data size. We observe that we
compare very well against Deep Speech 1 \& 2 which were trained with much
more data~\cite{hannun2014deep,amodei2015deep}.

Finally, we report in Table~\ref{tbl-best} the best results of our system
so far, trained on 1000h of speech, for each type of features. The overall
stride of architectures is 320 (see Figure~\ref{fig-raw-arch}), which
produces a label every 20 ms. We found that one could squeeze out about
$1\%$ in performance by refining the precision of the output. This is
efficiently achieved by shifting the input sequence, and feeding it to the
network several times. Results in Table~\ref{tbl-best} were obtained by a
single extra shift of 10 ms. Both power spectrum and raw features are
performing slightly worse than MFCCs. One could expect, however, that with
enough data (see Figure~\ref{fig-vs-h}) the gap would vanish.
\begin{table}
    \caption{LER/WER of the best sets of hyper-parameters for each feature types.}
    \label{tbl-best}
    \begin{center}
    \begin{tabular}{|l|cc|cc|cc|}
        \hline
         & \multicolumn{2}{|c|}{MFCC} & \multicolumn{2}{|c|}{PS} & \multicolumn{2}{|c|}{Raw} \\
         & LER & WER & LER & WER & LER & WER \\
        \hline
        dev-clean & 6.9 & & 9.3 & & 10.3 & \\
        test-clean & 6.9 & 7.2 & 9.1 & 9.4 & 10.6 & 10.1\\
        \hline
    \end{tabular}
    \end{center}
\end{table}

\section{Conclusion}

We have introduced a simple end-to-end automatic speech recognition system,
which combines a standard 1D convolutional neural network, a sequence criterion
which can infer the segmentation, and a simple beam-search decoder. The
decoding results are competitive on the LibriSpeech corpus with MFCC features
(7.2\% WER), and promising with power spectrum and raw speech (9.4\% WER and
10.1\% WER respectively). We showed that our AutoSegCriterion can be faster
than CTC \citep{graves2006connectionist}, and as accurate
(table~\ref{tbl-asg-vs-ctc}). Our approach breaks free from HMM/GMM
pre-training and force-alignment, as well as not being as computationally
intensive as RNN-based approaches \citep{amodei2015deep} (on average, one
LibriSpeech sentence is processed in less than 60ms by our ConvNet, and the
decoder runs at 8.6x on a single thread).

\bibliographystyle{acm} 
\bibliography{main}

\end{document}